\documentclass[runningheads]{llncs}
\usepackage{amsmath}
\usepackage{graphicx}
\usepackage{url}            % simple URL typesetting
\usepackage{booktabs}       % professional-quality tables
\usepackage{amsfonts}       % blackboard math symbols
\usepackage{nicefrac}       % compact symbols for 1/2, etc.
\usepackage{microtype}      % microtypography
\usepackage{amssymb}
\usepackage{wrapfig}
\usepackage{floatflt}
\usepackage{booktabs}
\usepackage{multirow}
\usepackage{setspace}
\usepackage{caption}
\usepackage{marvosym}

\begin{document}

\title{Exploring the Capabilities of Large Multimodal Models on Dense Text}
\titlerunning{Exploring the Capabilities of Large Multimodal Models on Dense Text}

\author{Shuo Zhang, Biao Yang, Zhang Li, Zhiyin Ma, Yuliang Liu(\Letter), Xiang Bai}
\authorrunning{S. Zhang et al.}

\institute{Huazhong University of Science and Technology \\
\email{\{u202014980,ylliu\}@hust.edu.cn}}

\maketitle               
%%%%%%%%%%%%%%%%%%%%%%%%%%%%%%%%%%%%%%%%%%%%%%%%%%%%%%%%%%%%%%%%%%%%%%%%%%%%%%%%%%
\begin{abstract}

While large multi-modal models (LMM) have shown notable progress in multi-modal tasks, their capabilities in tasks involving dense textual content remains to be fully explored.
Dense text, which carries important information, is often found in documents, tables, and product descriptions. Understanding dense text enables us to obtain more accurate information, assisting in making better decisions. To further explore the capabilities of LMM in complex text tasks, we propose the DT-VQA dataset, with 170k question-answer pairs.
In this paper, we conduct a comprehensive evaluation of GPT4V, Gemini, and various open-source LMMs on our dataset, revealing their strengths and weaknesses. Furthermore, we evaluate the effectiveness of two strategies for LMM: prompt engineering and downstream fine-tuning. We find that even with automatically labeled training datasets, significant improvements in model performance can be achieved. We hope that this research will promote the study of LMM in dense text tasks. Code will be released at \url{https://github.com/Yuliang-Liu/MultimodalOCR}.

\keywords{Large multi-modal model  \and Dense text visual question answering \and Evaluation.}
\end{abstract}

%%%%%%%%%%%%%%%%%%%%%%%%%%%%%%%%%%%%%%%%%%%%%%%%%%%%%%%%%%%%%%%%%%%%%%%%%%%%%%%%%%
\section{Introduction}

Dense text refers to the presence of a significant amount of textual information within images, such as documents and ingredient lists. Researching models' perception and understanding of text-dense images holds significant importance in various fields such as information extraction, text summary generation, smart search engines, and document analysis.

\begin{figure*}[h]
    \centering
    \includegraphics[width=0.8\linewidth]{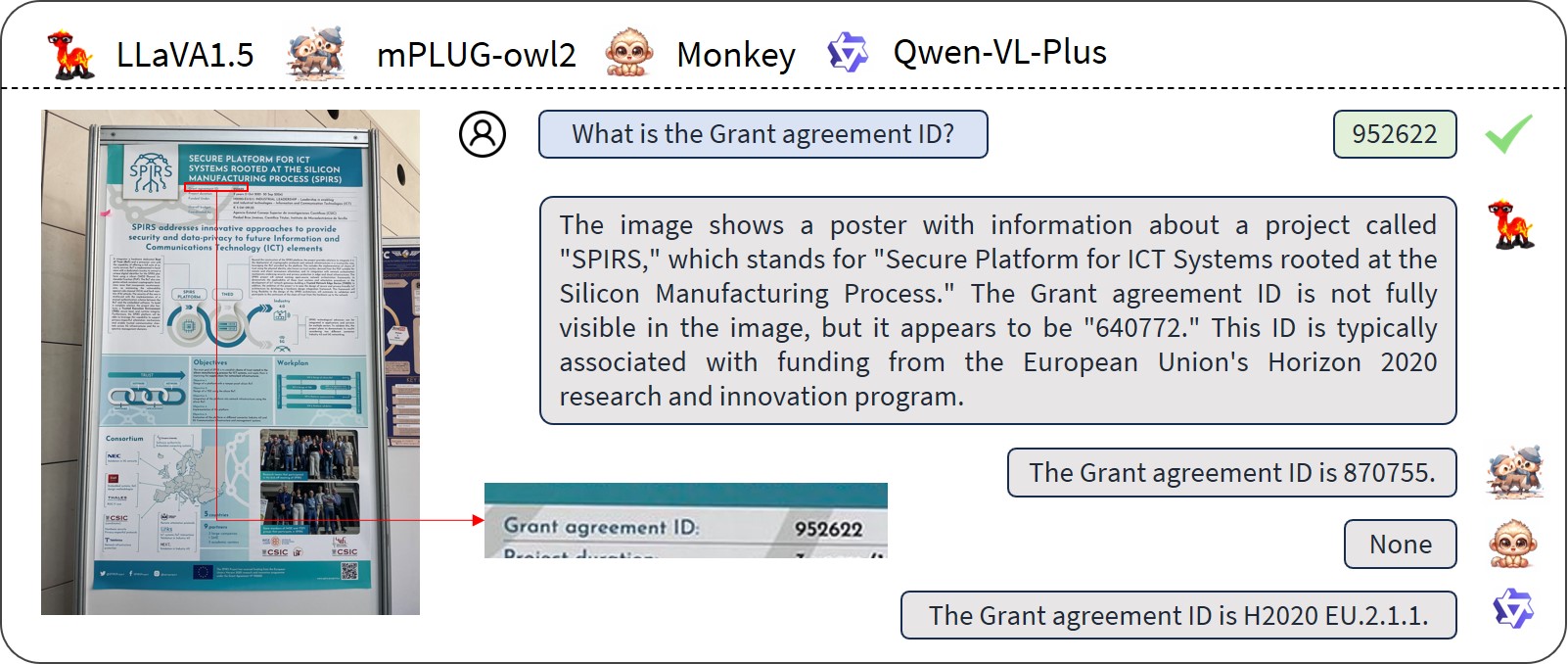}
    \caption{Visualization of VQA errors of LMMs on dense text images.}
    \label{error}
\end{figure*}

The field is rapidly evolving due to the capabilities of LMM in handling various types of data, such as images and text. 
Some LMMs~\cite{blip2,llava,instructblip,qwen-vl,llava1.5,monkey} have achieved success in various tasks, including image captioning and visual question answering (VQA), attracting academic interest. 
However, recent research~\cite{liu2023hidden} has demonstrated the limitations of LMMs in handling text-related visual tasks. For example, as shown in Fig.~\ref{error}, several LMMs fail to provide correct answers to questions about dense text images. This could be due to the low resolution of the input images and the lack of training, specifically on text images. 

To further evaluate the perceptual capabilities of LMMs for text, we propose a dataset named \textbf{D}ense \textbf{T}ext \textbf{V}isual \textbf{Q}uestion \textbf{A}nswering (DT-VQA), which covering outdoor scenes, tables, and product labels by collecting images and partial optical character recognition (OCR) information from HierText~\cite{hiertext}, DAST~\cite{dast}, POIE~\cite{poie}, and TabFact~\cite{tabfact}. 
Question-answer pairs are generated with the assistance of Gemini~\cite{gemini} and GPT-4V~\cite{gpt4v}. We carefully design the input prompts for Gemini~\cite{gemini} and GPT-4V~\cite{gpt4v} to seamlessly integrate images, OCR information, and instructions, guiding the generated question and answer pairs to be relevant to the text. Additionally, we constrain the response format to facilitate the extraction of question-answer pairs in post-processing. Ultimately, 170k question-answer pairs are obtained for 30k images.
For the test set, we manually select 1.2k question-answer pairs from 22k rather than choosing randomly to ensure diversity in the questions. Subsequently, we correct each question-answer pair in the test set and convert singular answers into answer lists, ensuring the correctness of the question-answer pairs and considering various possible correct answers.
The proposed dataset presents typical challenges due to: 1) \textit{High Information Density}, where texts are closely packed making word differentiation and boundary recognition difficult; 2) \textit{Complex Layout}, where images are with a wide range of layouts from unstructured scenes to semi-structured product images and structured tables; and 3) \textit{Various Text Types}, where images are with varied fonts, sizes, and arrangements, alongside diverse noises such as image blurring and uneven lighting.

To comprehensively evaluate the capabilities of existing LMMs, we perform extensive quantitative evaluation and qualitative analysis on the DT-VQA dataset. The experimental results highlight the significant challenges that even state-of-the-art LMMs encounter when faced with the complexity of processing dense text images.
To attempt to alleviate the limitations of LMMs in handling text, we try two solutions: prompt engineering and downstream fine-tuning. 
It has been noted that LMMs often generate responses that are longer and not directly related to the text present in the image. 
To address this issue, we design prompts before inputting questions, which significantly improved the performance of some models on the commonly used metric ANLS. In addition, we propose a new metric AccANLS to make up for the shortcomings of existing VQA task evaluations.
Furthermore, we fine-tune Qwen-VL~\cite{qwen-vl} and Monkey~\cite{monkey} using the DT-VQA training set, which consists of question-answer pairs automatically annotated by Gemini~\cite{gemini} and GPT-4V~\cite{gpt4v}, resulting in a significant improvement in performance on the test set. We summarize the advantages of our paper as follows:
\begin{itemize}
    \item We propose dense text-related visual question answering task, alongside with a carefully annotated dense text-related VQA dataset DT-VQA, to facilitate the research of the capabilities of LMMs for dense text.
    \item We conduct a quantitative evaluation and qualitative analysis on a wide range of LMMs. In response to the shortcomings of the existing evaluation metrics ANLS and accuracy, we propose a new evaluation metric called AccANLS. We find that models perform relatively better in tables and scene text images, where the boundaries between text are distinct, compared to product images. Currently, models face greater challenges in scenarios with denser text.
    \item To attempt to alleviate the limitations of LMMs in handling text, we propose two simple but effective strategies: prompt engineering and downstream fine-tuning. Prompt engineering brings an average improvement of 7.6\% on the ANLS metric for 13 LMMs, to some extent ensuring the effectiveness of ANLS metric evaluation; it also improves the overall performance of some models to a certain extent, bringing an average improvement of 0.7\% on the AccANLS evaluation metric. Downstream fine-tuning results in an improvement of 16.5\% and 12.2\% on the AccANLS evaluation metric for Qwen-VL and Monkey, respectively.
\end{itemize}

%%%%%%%%%%%%%%%%%%%%%%%%%%%%%%%%%%%%%%%%%%%%%%%%%%%%%%%%%%%%%%%%%%%%%%%%%%%%%%%%%%

\section{Related work}
\label{related work}

\subsection{Text-oriented Datasets}
Current datasets for text-rich images, including RCTW-17~\cite{rctw-17}, COCO-Text~\cite{coco-text}, Total-Text~\cite{total-text}, SynthText~\cite{synthtext}, CTW1500~\cite{ctw1500}, IC15~\cite{ic15}, MLT~\cite{mlt}, PubLayNet~\cite{publaynet}, and RVL-CDIP~\cite{RVL-CDIP} predominantly concentrate on foundational tasks such as text detection and layout analysis. However, these basic tasks are difficult to meet the needs of understanding and reasoning about text in images in practical applications. Antol~\cite{vqa} introduces an open VQA task, where the model is tasked with providing responses in natural language based on both image and question inputs. In contrast to traditional detection, recognition, and related tasks, VQA tasks impose elevated demands on the model's capacity to perceive and understand complex scenes. Existing datasets for VQA can be categorized as general and domain-specific, such as text-oriented or document-related, based on their application fields.

Text-oriented VQA datasets, such as OCR-VQA~\cite{ocrvqa}, TextVQA~\cite{textvqa}, ST-VQA~\cite{stvqa} and EST-VQA~\cite{estvqa}, require the model to answer questions by reading the text in the image. ChartQA~\cite{chartqa}, on the other hand, introduces intricate queries related to arithmetic and logical reasoning within the context of chart designs. Document-centric datasets like DocVQA~\cite{docvqa}, KLC~\cite{klc}, and VisualMRC~\cite{visualmrc} focus on images of documents, emphasizing structured layouts or containing tables. OCRBench~\cite{liu2023hidden} proposes a comprehensive evaluation system for various text-related visual tasks containing 29 related datasets. While these datasets acknowledge the significance of textual information within images, the density of text in these images is not sufficiently high. 
We introduce the DT-VQA dataset specifically designed to address the need for dense text, encompassing diverse styles such as scenes, product labels, and tables.

\subsection{Large Multi-modal Models}
In recent advancements within LMM research, notable efforts have been directed towards modal alignment and instruction fine-tuning. Flamingo~\cite{flamingo} uses gated cross attention to introduce visual features. LLaVA~\cite{llava} employs simple linear layers to map visual features to language space. BLIP2~\cite{blip2} uses multiple multi-modal pre-training tasks to guide the Q-Former module to map visual features to large language models(LLM). InstructBLIP~\cite{instructblip} integrates instructions into Q-Former, extracting task-related visual features. Qwen-VL~\cite{qwen-vl} introduces a position-aware visual language adapter that compresses image feature length. Monkey~\cite{monkey} proposes an efficient training method using sliding windows to improve resolution. 
In order to enable the model to better follow instructions and observe the image content in more detail, some work has also been done to enrich the data set. InstructBLIP~\cite{instructblip} converts 26 data sets from 11 tasks into an instruction-tuned format. LLaVAR~\cite{llavar} converts optical character recognition (OCR) results and image captions into plain text GPT-4~\cite{gpt-4}, creating a rich instruction tracking dataset. Despite the various strides made by existing LMMs in different dimensions, their proficiency in dense text-related image understanding still faces limitations due to the absence of high-quality datasets dedicated to this specific domain. Our newly introduced DT-VQA dataset endeavors to bridge this gap by providing a valuable dataset designed for answering questions related to dense text.

%%%%%%%%%%%%%%%%%%%%%%%%%%%%%%%%%%%%%%%%%%%%%%%%%%%%%%%%%%%%%%%%%%%%%%%%%%%%%%%%%%

\section{DT-VQA Dataset}
Dense text images are highly effective at conveying substantial amounts of information within a constrained space.
To further explore the text-aware capabilities of LMMs, we propose \textbf{D}ense \textbf{T}ext \textbf{V}isual \textbf{Q}uestion \textbf{A}nswering dataset (DT-VQA). We select a diverse array of images spanning scenes, tables, and product labels sourced from multiple datasets. 
We meticulously design prompts to feed images, partially annotated OCR data, and specific-format question-answer pair generation instructions into Gemini~\cite{gemini} and GPT-4V~\cite{gpt4v}, resulting in the creation of 170k question-answer pairs from a dataset of 30k images.
We perform random checks on the generated question-answer pairs to ensure their quality.
For the test set, we manually screen and correct the question-answer pairs and convert single answers into answer lists, obtaining the final 1.2k question-answer pairs from the original pool of 22k.

\subsection{Images}
Current text-related datasets mainly focus on scene images, or single-source documents or tables, limiting the diversity of image styles, which hinders the comprehensiveness of model training and evaluation. Additionally, these images have lower text density. To address this issue, we aim to construct a dataset that focuses on dense text and exhibits diverse styles. Therefore, we collect images from four text-rich datasets of different styles, which include clearly structured tables, semi-structured labels, and unstructured scene images, enabling exploration of model adaptability across various text-related application scenarios.

\textbf{Hiertext}~\cite{hiertext} is a scene image dataset with multi-level annotations, introduced to facilitate the convergence of scene text detection and layout analysis tasks. The images in HierText are sourced from the Open Images~\cite{openimagev4} dataset, where OCR tools are employed to identify text-containing images. Images with sparse words, low recognition confidence, and non-English dominant text are filtered out. The resulting dataset comprises scene images with high-density text, averaging over 100 words per image. The text information in HierText is annotated using a multi-level approach, initially marking word positions with polygons and subsequently grouping words into text lines.

\textbf{POIE}~\cite{poie}, a product-oriented dataset tailored for Visual Information Extraction (VIE) tasks, comprises camera images capturing nutritional labels on various products. The dataset exhibits a range of challenges, as the images showcase different layouts with varying degrees of folding, bending, deformation, and perspective distortion. The inherent complexities in POIE contribute to a more intricate and realistic representation of real-world scenarios. These nuances make this dataset particularly valuable for evaluating and enhancing the robustness and generalization capabilities of VIE models in preparation for more demanding real-world applications.

\textbf{DAST}~\cite{dast} is a dataset focused on dense and irregularly shaped scene text in commodity images. It predominantly includes images of products with intricate details presented on small, crinkled packaging, sourced from the internet. Notably, it incorporates Chinese product labels. Given that product labels tend to be text-dense, coupled with variations in camera angles and the curvature of the product packaging, the text often appears curved and assumes diverse shapes, presenting additional challenges for analysis.

\textbf{TabFact}~\cite{tabfact} is a dataset focused on fact verification under structured evidence. It contains 16k Wikipedia tables, covering topics like game records, data statistics, and more, so it may involve simple numerical logic and reasoning, which brings certain challenges to the model.
We also observe that these well-organized tables contain abundant layout information, which is crucial for evaluating the model's capabilities in text positioning, understanding, and reasoning.

The test set images of DT-VQA come from the test sets of the above four datasets to ensure that no data leakage occurs in the LMMs evaluated later.

\subsection{Annotation}
\begin{figure*}[h]
    \centering
    \includegraphics[width=0.8\linewidth]{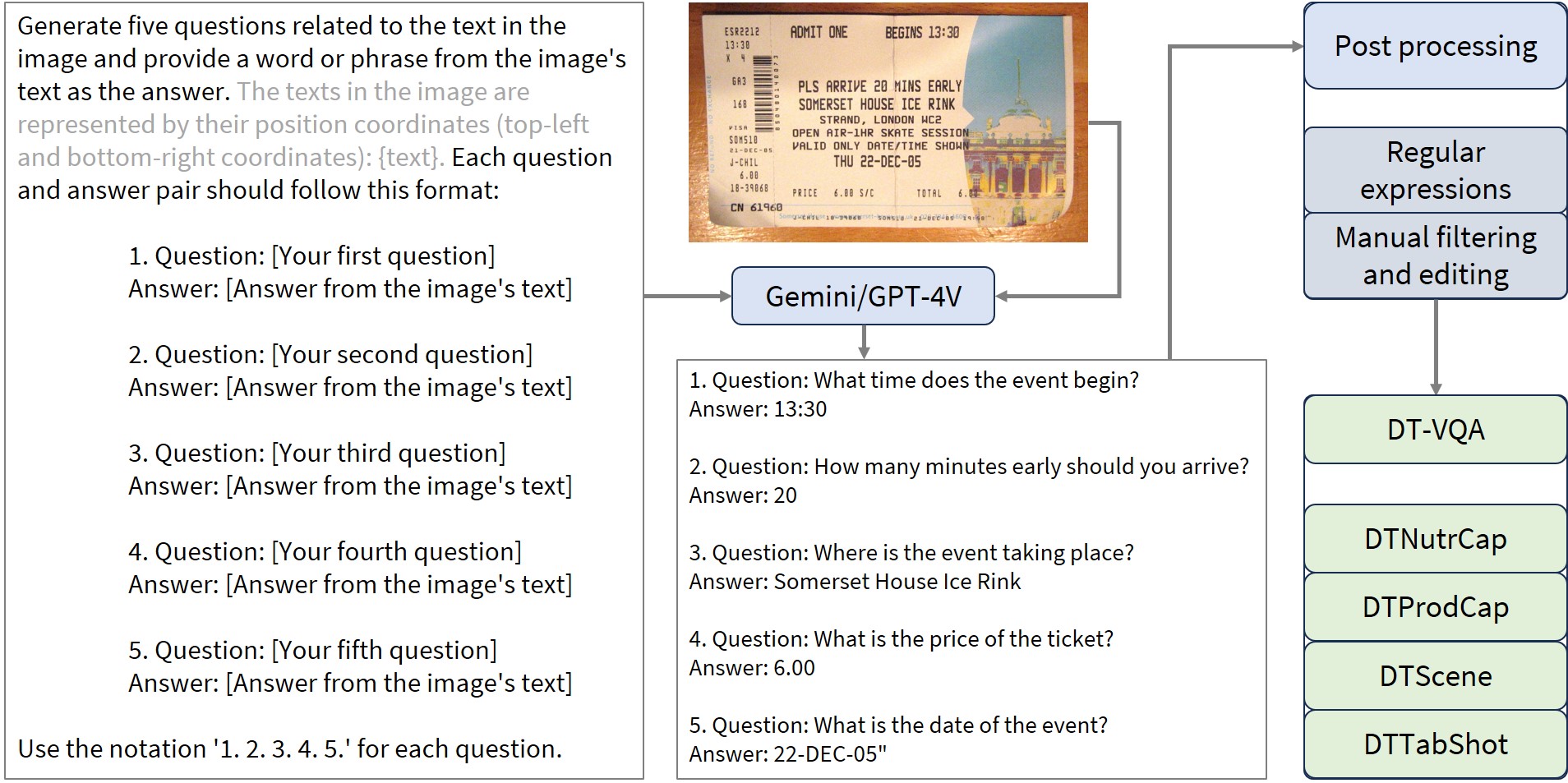}
    \caption{Question-answer pair generation pipeline. (Note: Only the question and answer pairs from Hiertext~\cite{hiertext} and POIE~\cite{poie} are generated with OCR information input, which is the gray part in the picture above.)}
    \label{pipeline}
\end{figure*}

The generation of question-answer pairs is assisted by Gemini~\cite{gemini} and GPT-4V~\cite{gpt4v}. We feed images, OCR information and instructions guiding the generation into these two LMMs, producing five question-answer pairs for each image. Subsequently, we organize and manually filter the obtained question-answer pairs in test set. The pipeline is shown in Fig.~\ref{pipeline}.

\textbf{Instruction designing.} Due to the diverse sources of images and varying annotation methods, we broadly categorize the images into two groups: those annotated with textual content (Hiertext~\cite{hiertext} and POIE~\cite{poie}) and those without textual content annotations (DAST~\cite{dast} and TabFact~\cite{tabfact}). 
For images without text annotation, we only use the image to guide Gemini~\cite{gemini} to perform the process of generating question and answer pairs.
For images annotated with text, we maximize the utilization of existing annotations. We feed text and location information to Gemini~\cite{gemini} to generate more text-related question-answer pairs. Accurate text input also enhances the precision of generated answers, reducing the probability of low-quality question-answer pairs due to potential text recognition errors in Gemini~\cite{gemini} and GPT-4V~\cite{gpt4v}.
The phrase "[Answer from the image's text]" in the prompt guides the model to generate text-related questions and extract answers from the image's text.
In addition, the detailed generation format we give is also conducive to extracting question and answer pairs in post-processing.
For all images from the four datasets, we use Gemini~\cite{gemini} to generate five question-answer pairs each. Specifically, due to the complexity of scene images, we additionally use GPT-4V~\cite{gpt4v} for another round of question-answer pair generation for the training set of HierText~\cite{hiertext}, which consists of 8k images. Detailed statistical information on the DT-VQA training set is shown in Tab.~\ref{DT-VQA_train}.

\begin{table*}[]
\centering
\caption{DT-VQA training set question and answer statistics. DTNutrCap, DTProdCap, DTScene, and DTTabShot represent datasets generated from POIE's nutritional label photos, DAST's product photos, HierText's scene text images, and TabFact's Wikipedia table screenshots, respectively. (Only images from the Hiertext training set use GPT-4V to generate question and answer pairs, and images from the training set and validation set of the four data sets use Gemini to generate question and answer pairs.)}
\label{DT-VQA_train}
\scalebox{1}{
\begin{tabular}{l|cccc}
\toprule
Subset                     & Source                    & Tool   & QA Pairs/Image & Number of Images\\ \midrule
DTNutrCap                  & POIE~\cite{poie}                      & Gemini & 5              & 3000\\
DTProdCap                  & DAST~\cite{dast}                      & Gemini & 5              & 1538\\
\multirow{2}{*}{DTScene}   & \multirow{2}{*}{Hiertext~\cite{hiertext}} & Gemini & 5              & 10005 \\
                           &                           & GPT-4V & 5              & 8281\\
DTTabShot                  & TabFact~\cite{tabfact}                   & Gemini & 5              & 14127 \\ \bottomrule
\end{tabular}}
\end{table*}

\textbf{Post processing.} For the initial response generated by Gemini~\cite{gemini} and GPT-4V~\cite{gpt4v}, we first use regular expressions to extract question-answer pairs. Additionally, all test set images lack original text annotations. To ensure the quality of the test set question-answer pairs, we conduct manual screening and select 300 high-quality question-answer pairs for each style of image. 
For answers in the test set that raised doubts, we conduct further processing. For instance, considering the answer ``2g,'' variations like ``2 g,'' ``2 grams,'' and ``2'' can be considered correct. Therefore, we manually transform the original single answer into a list of potential answers one by one. For each evaluation metric, we select the score of the highest-ranking answer from this list. 

\subsection{Statistics and Analysis}

\begin{figure*}[h]
    \centering
    \includegraphics[width=0.8\linewidth]{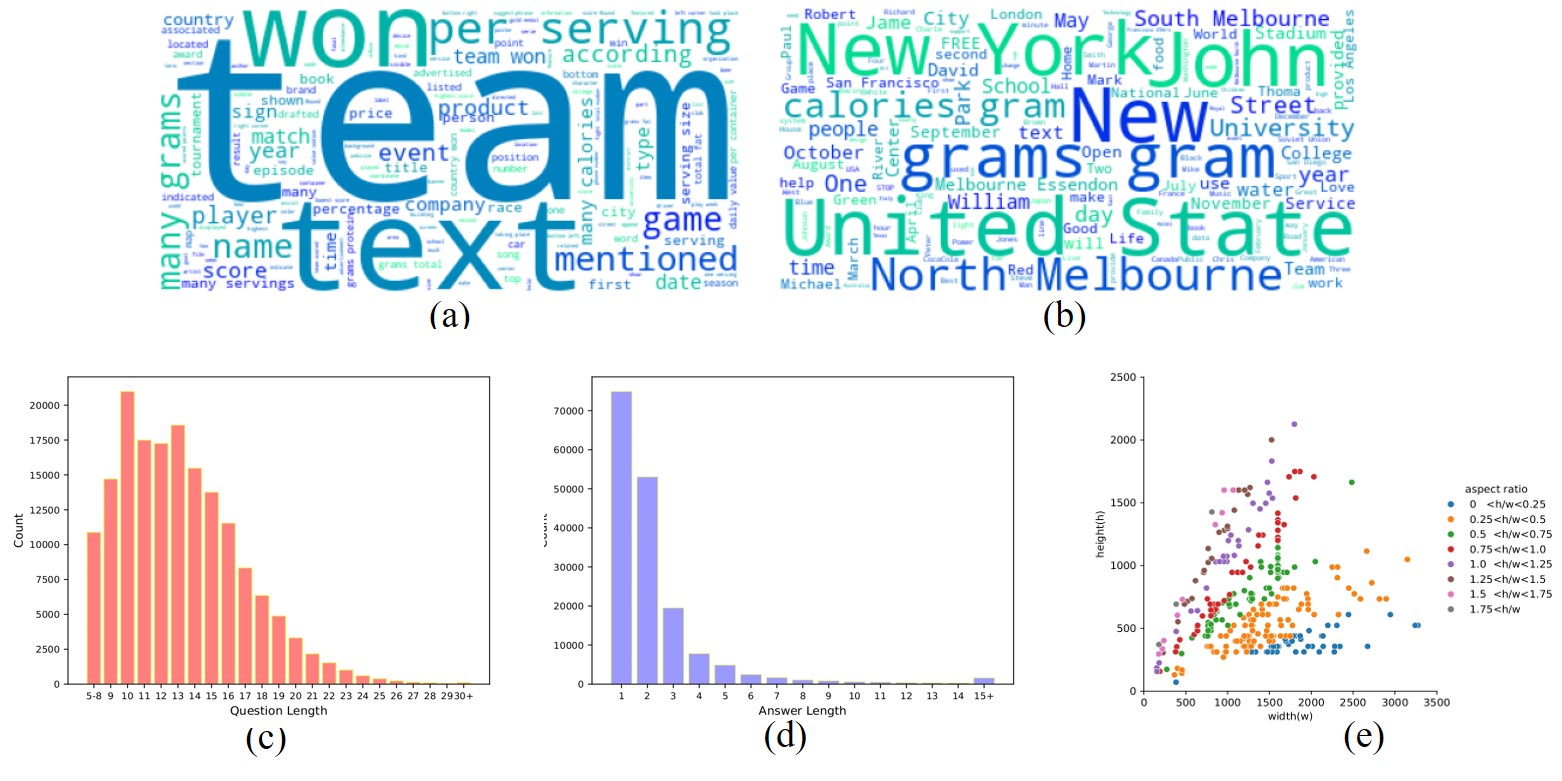}
    \caption{Statistics and analysis of DT-VQA.}
    \label{data}
\end{figure*}

We have separately analyzed the frequency of words in the questions and answers of the dataset and presented them in the form of word clouds. Fig.~\ref{data}(a) corresponds to the frequency of different words in the questions, revealing that the word "team" has the highest occurrence in questions. This is attributed to the TabFact dataset predominantly recording information about various teams, causing the word ``team'' to appear frequently in questions generated by Gemini for TabFact. Meanwhile, Fig.~\ref{data}(b) represents the frequency of different words in the answers, where high-frequency words mostly include place names or unit names, showing a relatively even distribution and indicating good diversity in response to different questions.

In Fig.~\ref{data}(c) and (d), we illustrate the distribution of question and answer lengths in our dataset. The horizontal axis represents the number of words in questions or answers, while the vertical axis denotes the frequency of occurrences at each length. Questions typically consist of at least five words, with the majority falling within the range of 10 to 15 words. Answer lengths are concentrated in the range of 1 to 3 words, emphasizing concise responses. In rare instances, Gemini-generated answers may exceed 20 words. This analysis suggests a reasonable distribution of question and answer lengths in our dataset, aligning well with the requirements of dense text-based question answering.

Finally, Fig.~\ref{data}(e) randomly selects 400 images from all used images and presents their dimensions in a scatter plot. The aspect ratios of the images fall within a reasonable range, with no occurrences of excessively wide or tall images. And there are occasionally images with a large aspect ratio that increase the diversity of the data.

%%%%%%%%%%%%%%%%%%%%%%%%%%%%%%%%%%%%%%%%%%%%%%%%%%%%%%%%%%%%%%%%%%%%%%%%%%%%%%%%%%
\section{Benchmark}
\subsection{Evaluation Metric}
Unlike the deterministic scoring of multiple-choice answers, assessing the accuracy of open-ended question responses presents certain challenges. We use three metrics to score the answers.

\textbf{ANLS} (Average Normalized Levenshtein Similarity)~\cite{stvqa} can apply slight penalties in situations where the expected response is correct but there is a slight deviation in text recognition. We use the default threshold of 0.5 during the evaluation process, indicating the output of the metric is 0 if its value is lower than 0.5. 
This threshold helps us figure out if the answer was right but not properly recognized, or if the wrong text from the image was chosen as the answer.
However, when LMMs cannot accurately determine the expected length of the generated answers, they may also produce excessively long answers in VQA tasks, resulting in ANLS results no longer effectively reflecting the correctness of the answers and the performance of the models.

\textbf{Accuracy} serves as our second evaluation criterion. Since LMMs often generate long response, which may contain the correct answer but result in low ANLS scores due to significant differences in length compared to concise standard answers, we use accuracy as an alternative method. It evaluates if the correct answer appears in the output; when present, the response is deemed accurate.

\textbf{AccANLS} is a novel evaluation criterion that we have newly introduced. 

To balance the issues arising from text recognition errors and excessively long output lengths, we utilize AccANLS as a comprehensive assessment score. For a given question-answer pair, we initially check if the accuracy is 1; if not, we calculate the score using the ANLS method. The calculation formula is as follows:

\begin{equation}
\begin{aligned}
\text{AccANLS} &= \frac{1}{N}\sum_{i=0}^{N}\left(\max_j h(a_{ij}, o_{q_i})\right), \\
h(a_{ij}, o_{q_i}) &=
\begin{cases}
1 & \text{if } a_{ij} \text{ is within } o_{q_i} \\
f(a_{ij}, o_{q_i}) & \text{otherwise}
\end{cases},\\
f(a_{ij}, o_{q_i}) &=
\begin{cases}
(1 - NL(a_{ij}, o_{q_i})) & \text{if}~NL(a_{ij}, o_{q_i}) < \tau \\
0 & \text{if}~NL(a_{ij}, o_{q_i}) \geq \tau
\end{cases},
\end{aligned}
\end{equation}
where i represents the $i_{th}$ question, j represents the $j_{th}$ reference answer provided, NL calculates the normalized Levenshtein distance between the strings $a_{ij}$ and $o_{q_i}$, and $\tau$ is the selected threshold.

\subsection{Baseline}

OCRBench~\cite{liu2023hidden} provides a comprehensive examination of the abilities of LMMs in the field of OCR. In line with this study, we have chosen several widely used general-purpose LMMs, along with some models specifically designed for text-related tasks, to further investigate the potential of LMMs in processing dense text.

LLaVA~\cite{llava} utilizes a linear layer to connect CLIP's~\cite{clip} open-set visual encoder with LLaMA~\cite{llama}. Besides, it explores an efficient training approach for pretraining and fine-tuning a visual language model. 
LLaVA1.5~\cite{llava1.5} replaces the linear layer with a two-layer MLP, adopts a higher resolution CLIP visual encoder, and leverages richer data, resulting in a more robust baseline.
LLaVAR~\cite{llavar} follows LLaVA's model structure and two-stage training paradigm. To better comprehend text details in images, it introduces OCR results and image captions as prompts to the pure-text GPT-4~\cite{gpt-4}. This creates a rich instruction tracking dataset added to the training set, significantly enhancing LLaVA's performance on text-based VQA datasets.
BLIP-2~\cite{blip2} employs multiple multi-modal pretraining tasks to guide the Q-Former module in mapping visual features to the LLM. 
InstructBLIP~\cite{instructblip}, building on BLIP2's model structure, inputs instructions to the Q-Former, extracting task-relevant visual features. Additionally, it transforms 26 datasets (11 tasks) into instruction-tuning format to enrich the data set. 
BLIVA~\cite{bliva} utilizes the Q-former structure from InstructBLIP and the LLaVA approach of directly inputting visual tokens through linear projection to the LLM. It significantly improves VQA capabilities in text-rich scenes.
mPLUG-owl~\cite{mplug-owl} introduces a novel training approach by unfreezing the visual encoder and jointly adjusting the language module's LoRA~\cite{lora} parameters using multimodal and unimodal data. 
mPLUG-owl2~\cite{mplug-owl2} integrates shared functional modules to promote modal collaboration and introduces a modal adaptive module to retain module-specific functionality. It is the first LMM to simultaneously demonstrate modal collaboration phenomena in pure-text and multi-modal scenarios.
Qwen-VL~\cite{qwen-vl} introduces a position-aware visual language adapter that compresses image feature lengths, raising the input resolution to 448 for the first time. 
Monkey~\cite{monkey} builds upon Qwen-VL with an efficient training method using sliding windows to further increase resolution and generate a dataset with detailed descriptions. Higher resolution and more detailed description data contribute to better capturing details in images, such as text information.

\subsection{Strategy}

\begin{figure*}[h]
    \centering
    \includegraphics[width=0.8\linewidth]{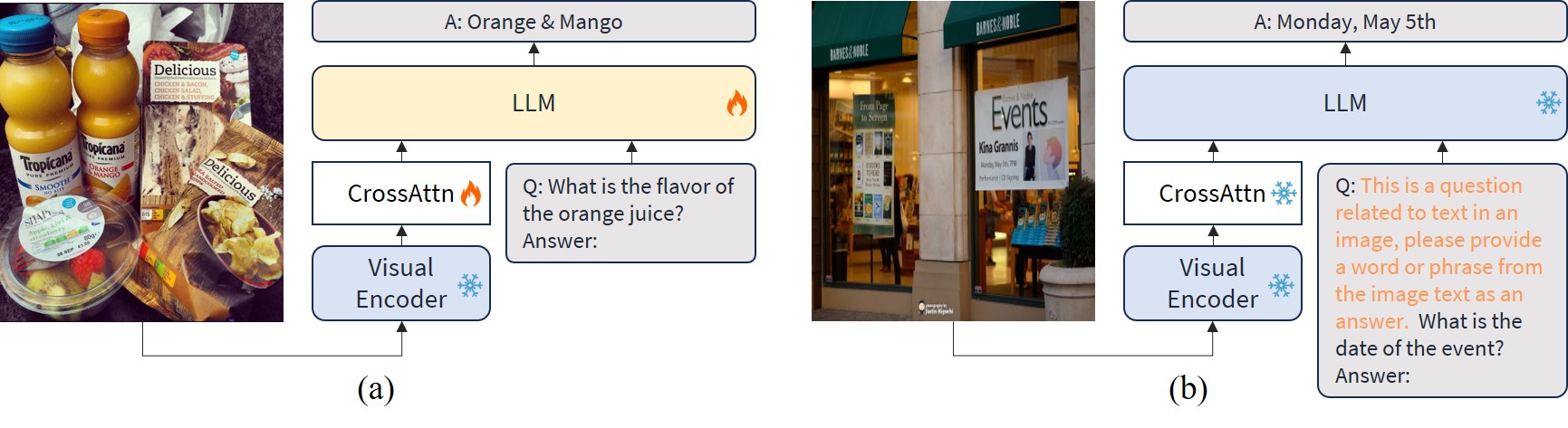}
    \caption{(a) Downstream fine-tuning. (b) Prompt engineering.}
    \label{prompt_finetune_model}
\end{figure*}

We evaluate the effectiveness of two strategies for LMM: prompt engineering and downstream fine-tuning. Taking Qwen-VL~\cite{qwen-vl} as an example, we illustrate the specific implementation of these two strategies in the Fig.~\ref{prompt_finetune_model}.

\textbf{Prompt engineering,} Given the ultimate goal of application in various daily scenarios, LMMs are typically trained on diverse tasks like image captioning and various VQA formats. During training, specific instructions tailored to different expected answer lengths are provided. Consequently, in the absence of additional prompts, LMMs cannot autonomously determine the expected answer length accurately. As mentioned in Section 4.1, LMMs often produce excessively long answers during VQA tasks, rendering the ANLS evaluation method ineffective. To address this issue, we have introduced an additional prompt text at the beginning of the question: ``This is a question related to text in an image, please provide a word or phrase from the image text as an answer.'' This modification enables the model to generate shorter format answers and focus more on the textual information within the image.

\textbf{Fine-tuning.} Considering the challenges posed by dense text datasets, we conduct fine-tuning on Qwen-VL~\cite{qwen-vl} and Monkey~\cite{monkey} using the DT-VQA dataset.
Given that images from DAST include a significant number of Chinese images, and the Gemini-generated question-answering pairs contain a mix of Chinese and English, which may not be conducive to training the LMM to correctly extract text from images to answer questions, we exclude this portion of training data, \textit{i.e.}, we use question-answering pairs in DTScene, DTNutrCap, and DTTabShot for fine-tuning.
We adopt the model scale configurations of Qwen-VL~\cite{qwen-vl} and Monkey~\cite{monkey}. Throughout the training process, we utilize the AdamW optimizer~\cite{adamw} with a learning rate set to 5e-6 and follow a cosine learning rate schedule. We set the values of $\beta_1$ and $\beta_2$ to 0.9 and 0.95, respectively. Additionally, we apply a warm-up period of 100 steps, utilize a batch size of 128 and apply weight decay of 0.1.

%%%%%%%%%%%%%%%%%%%%%%%%%%%%%%%%%%%%%%%%%%%%%%%%%%%%%%%%%%%%%%%%%%%%%%%%%%%%%%%%%%
\section{Result}
\subsection{Quantitative Result}

\begin{table*}[h]
\centering
\caption{Results of LMMs on DT-VQA.}
\label{result}
\resizebox{1\columnwidth}{!}{
\begin{tabular}{l|ccc|ccc|ccc|ccc|c}
\toprule
\multirow{2}{*}{model}   & \multicolumn{3}{c|}{DTNutrCap} & \multicolumn{3}{c|}{DTProdCap} & \multicolumn{3}{c|}{DTScene} & \multicolumn{3}{c|}{DTTabShot} & Avg \\
                         & ANLS   & Accuracy   & AccANLS  & ANLS   & Accuracy   & AccANLS  & ANLS  & Accuracy  & AccANLS  & ANLS   & Accuracy  & AccANLS  & AccANLS \\ \midrule
BLIP2~\cite{blip2}        & 0.5    & 2.3        & 2.9      & 7.3    & 9.7        & 13.0     & 15.6  & 11.3      & 17.6     & 10.3   & 7.3       & 11.0     & 11.1 \\
LLaVA\_7b~\cite{llava}                  & 0.0    & 9.3        & 9.3      & 0.0    & 10.0       & 10.0     & 0.0   & 12.7      & 12.7     & 0.0    & 16.0      & 16.0     & 12.0 \\
InstructBLIP~\cite{instructblip} & 4.8    & 7.7        & 8.6      & 5.1    & 10.7       & 12.1     & 10.2  & 15.7      & 19.7     & 11.4   & 11.0      & 13.0     & 13.4 \\
BLIVA~\cite{bliva}                    & 0.4    & 10.0       & 10.4     & 1.5    & 11.7       & 12.1     & 1.5   & 20.0      & 21.2     & 0.0    & 13.0      & 13.0     & 14.2 \\
mPLUG-owl~\cite{mplug-owl}                & 0.0    & 9.0        & 9.0      & 0.0    & 17.7       & 17.7     & 0.2   & 21.3      & 21.5     & 0.0    & 16.3      & 16.3     & 16.1 \\
LLaVAR~\cite{llavar}                   & 0.0    & 17.7       & 17.7     & 0.2    & 23.0       & 23.2     & 0.4   & 26.7      & 26.8     & 0.2    & 24.3      & 24.3     & 23.0\\
LLaVA1.5\_7b~\cite{llava1.5}            & 13.3   & 10.0       & 14.9     & 25.6   & 14.7       & 26.8     & 30.7  & 19.3      & 30.8     & 26.5   & 20.0      & 27.0     & 24.9\\
mPLUG-owl2~\cite{mplug-owl2}               & 0.0    & 24.3       & 24.3     & 0.2    & 27.3       & 27.5     & 0.6   & 33.3      & 33.9     & 0.0    & 23.0      & 23.0     & 27.2\\
LLaVA1.5\_13b~\cite{llava1.5}           & 23.0   & 13.7       & 25.9     & 29.9   & 18.7       & 30.7     & 34.0  & 18.7      & 34.2     & 29.7   & 23.3      & 29.9     & 30.2 \\
Qwen-VL~\cite{qwen-vl}                   & 48.7   & 37.0       & 50.2     & 52.7   & 44.0       & 55.9     & 50.6  & 40.0      & 52.2     & 47.6   & 45.3      & 49.3     & 51.9 \\
Monkey~\cite{monkey}                   & \textbf{51.0}   & 39.7       & 52.1     & \textbf{64.6}   & 56.0       & 68.0     & \textbf{58.0}  & 45.7      & 58.9     & \textbf{48.5}   & 43.3      & 49.4     & 57.1 \\
GPT-4V~\cite{gpt4v}                    & 0.0    & \textbf{74.7}       & \textbf{74.7}     & 0.0    & 54.3       & 54.3     & 0.0   & 68.0      & 68.0     & 0.0    & \textbf{88.3}      & \textbf{88.3}     & 71.3 \\
Gemini~\cite{gemini}                   & 24.1   & 68.3       & 69.4     & 30.7   & \textbf{65.7}       & \textbf{68.3}     & 31.2  & \textbf{70.7}      & \textbf{74.7}     & 11.2   & 82.7      & 82.9     & \textbf{73.8} \\ \bottomrule
\end{tabular}}
\end{table*}

\begin{table*}[h]
\centering
\caption{Results of LMMs with prompt input on DT-VQA.}
\label{result_prompt}
\resizebox{1\columnwidth}{!}{
\begin{tabular}{l|ccc|ccc|ccc|ccc|c}
\toprule
\multirow{2}{*}{model}   & \multicolumn{3}{c|}{DTNutrCap} & \multicolumn{3}{c|}{DTProdCap} & \multicolumn{3}{c|}{DTScene} & \multicolumn{3}{c|}{DTTabShot}  & Avg \\
                         & ANLS   & Accuracy  & AccANLS  & ANLS   & Accuracy  & AccANLS  & ANLS  & Accuracy  & AccANLS & ANLS   & Accuracy   & AccANLS & AccANLS \\ \midrule
BLIP2~\cite{blip2}                  & 1.2    & 1.7       & 1.9      & 10.0   & 6.7       & 11.1     & 15.9  & 12.3      & 17.8    & 8.7    & 6.7       & 9.6      & 10.1 \\
LLaVA\_7b~\cite{llava}                & 0.0    & 9.3       & 9.3      & 0.0    & 10.0      & 10.0     & 0.0   & 11.0      & 11.0    & 0.0    & 12.7      & 12.7     & 10.8 \\
BLIVA~\cite{bliva}                  & 0.0    & 11.7      & 11.7     & 0.0    & 13.3      & 13.3     & 0.3   & 16.3      & 16.3    & 0.0    & 13.0      & 13.0     & 13.6 \\
InstructBLIP~\cite{instructblip}           & 5.4    & 5.3       & 5.8      & 22.2   & 10.3      & 22.5     & 17.3  & 11.7      & 17.9    & 8.3    & 8.0       & 9.2      & 13.9 \\
mPLUG-owl~\cite{mplug-owl}              & 0.0    & 10.3      & 10.3     & 0.0    & 19.0      & 19.0     & 0.2   & 21.0      & 21.2    & 0.0    & 16.7      & 16.7     & 16.8 \\
LLaVAR~\cite{llavar}                 & 0.0    & 17.3      & 17.3     & 0.0    & 22.7      & 22.7     & 0.6   & 25.7      & 25.8    & 0.2    & 23.3      & 23.3     & 22.3 \\
mPLUG-owl2~\cite{mplug-owl2}             & 3.7    & 20.0      & 20.4     & 12.3   & 26.7      & 29.5     & 16.2  & 25.0      & 30.7    & 8.9    & 20.3      & 21.8     & 25.6 \\
LLaVA1.5\_7b~\cite{llava1.5}           & 15.0   & 12.0      & 16.7     & 29.2   & 17.3      & 30.8     & 33.9  & 21.0      & 34.0    & 25.6   & 18.0      & 25.9     & 26.9 \\
LLaVA1.5\_13b~\cite{llava1.5}          & 23.8   & 15.0      & 26.3     & 32.1   & 19.7      & 33.3     & 33.4  & 18.3      & 33.6    & 28.2   & 22.7      & 28.5     & 30.4 \\
Qwen-VL~\cite{qwen-vl}                 & 47.0   & 35.3      & 48.5     & 57.8   & 48.0      & 61.0     & 53.2  & 45.0      & 55.1    & 44.7   & 41.0      & 45.8     & 52.6 \\
Monkey~\cite{monkey}                 & 49.5   & 38.0      & 51.2     & \textbf{65.6}   & 58.0      & 67.9     & \textbf{58.3}  & 45.3      & 59.0    & 46.5   & 44.0      & 48.2     & 56.6 \\
GPT-4V~\cite{gpt4v}                  & 65.7   & \textbf{78.7}      & \textbf{86.1}     & 22.3   & 57.3      & 59.7     & 23.4  & 66.0      & 71.0    & \textbf{63.5}   & \textbf{88.7}      & \textbf{90.3}     & 76.8 \\
Gemini~\cite{gemini}                 & \textbf{78.1}   & 75.3      & 82.5     & 60.3   & \textbf{70.3}      & \textbf{74.5}     & 57.9  & \textbf{72.7}      & \textbf{78.6}    & 47.8   & 80.7      & 81.9     & \textbf{79.4} \\ \bottomrule    
\end{tabular}}
\end{table*}

The results obtained by inputting images and questions into LMMs are presented in Tab.~\ref{result}, while the results with additional prompts mentioned in Section 4.3 are shown in Tab.~\ref{result_prompt}. According to the experimental results, we have the following findings:
\begin{itemize}
    \item The average scores of LMMs on ANLS are significantly lower than Accuracy, indicating a notable issue of generating overly lengthy outputs. The AccANLS metric, which considers both text misrecognition and output length issues, provides a more comprehensive reflection of the model's capabilities compared to individual ANLS or accuracy metrics. We use AccANLS across the entire test set for ranking.
    \item For some models, the addition of prompts results in a noticeable increase in ANLS scores, such as InstructBLIP~\cite{instructblip} and GPT-4V~\cite{gpt4v}, indicating a strong adherence to instructions. Some models, such as Monkey~\cite{monkey}, have slightly lower scores on AccANLS after adding additional prompts. This may be because the model has paid enough attention to the text part during training and no longer needs additional prompts. Designing prompts that are more suitable for each model is a problem that needs further exploration.
    % Models with minimal changes in ANLS scores may struggle with handling dense text or inherently produce short and text-focused outputs.
    \item The scores of LMMs on DTNutrCap and DTProdCap are relatively lower compared to scores on DTScene and DTTabShot, suggesting that the models perform relatively well in scenarios with sparse text in tables and scene text images but face challenges with images with too compact and dense text.
    \item Excluding GPT-4V~\cite{gpt4v} and Gemini~\cite{gemini}, among open-source models, Qwen-VL~\cite{qwen-vl} and Monkey~\cite{monkey} demonstrate the best performance. This indicates to some extent that improving the input image resolution significantly contributes to the recognition and understanding of dense text images.
    \item Considering the assistance provided by Gemini~\cite{gemini} in generating question-answer pairs for the test set, it is expected to achieve a high score. Additionally, even with an additional prompt, GPT-4V~\cite{gpt4v} only achieves a score of 76.8. Among open-source models, Monkey~\cite{monkey}, the top-performing model, scores only 56.6. The performance of LMMs still has significant room for improvement, affirming the importance of introducing datasets specifically tailored for dense text question-answering.
\end{itemize}

We further explore the models Qwen-VL~\cite{qwen-vl} and Monkey~\cite{monkey}, which demonstrated relatively better performance. We use prompt engineering and downstream fine-tuning and the results are recorded in Tab.~\ref{ablation}. Simply adding prompts to the questions did not show significant improvements on these two models, possibly because the models themselves can discern the DT-VQA task's requirement of generating short answers from textual content in images. However, after fine-tuning on the training set of DT-VQA, both models exhibit substantial improvement, indicating that even with data labeled automatically, models performance can be significantly enhanced.

\begin{table*}[]\scriptsize
\centering
\caption{Results of prompt engineering and downstream fine-tuning  experiments on Qwen-VL~\cite{qwen-vl} and Monkey~\cite{monkey}.}
\label{ablation}
\resizebox{1\columnwidth}{!}{
\begin{tabular}{l|ccc|ccc|ccc|ccc|c}
\toprule
\multirow{2}{*}{model} & \multicolumn{3}{c|}{DTNutrCap}           & \multicolumn{3}{c|}{DTProdCap}           & \multicolumn{3}{c|}{DTScene}             & \multicolumn{3}{c|}{DTTabShot}                 & Avg \\
                       & ANLS   & Accuracy  & AccANLS  & ANLS   & Accuracy  & AccANLS  & ANLS  & Accuracy  & AccANLS & ANLS   & Accuracy  & AccANLS   & AccANLS\\ \midrule
Qwen-VL~\cite{qwen-vl}                 & 48.7   & 37        & 50.2     & 52.7   & 44        & 55.9     & 50.6  & 40        & 52.2    & 47.6   & 45.3      & 49.3     & 51.9 \\
+prompt         & 47.0   & 35.3      & 48.5     & 57.8   & 48.0      & 61.0     & 53.2  & 45.0      & 55.1    & 44.7   & 41.0      & 45.8     & 52.6 \\
+finutune       & 65.8   & 52.7      & 66.7     & 63.3   & 59.0      & 67.6     & 69.0  & 57.3      & 71.3    & 66.7   & 65.3      & 67.8     & 68.4 \\ \midrule
Monkey~\cite{monkey}                 & 51     & 39.7      & 52.1     & 64.6   & 56        & 68       & 58    & 45.7      & 58.9    & 48.5   & 43.3      & 49.4     & 57.1 \\
+prompt         & 49.5   & 38.0      & 51.2     & 65.6   & 58.0      & 67.9     & 58.3  & 45.3      & 59.0    & 46.5   & 44.0      & 48.2     & 56.6 \\
+finutune       & 66.9   & 55.3      & 68.2     & 63.7   & 57.7      & 68.1     & 69.1  & 60.0      & 71.8    & 67.9   & 65.3      & 69.0     & 69.3 \\ \bottomrule
\end{tabular}}
\end{table*}

\subsection{Qualitative Result}

\begin{figure*}[h]
    \centering
    \includegraphics[width=0.8\linewidth]{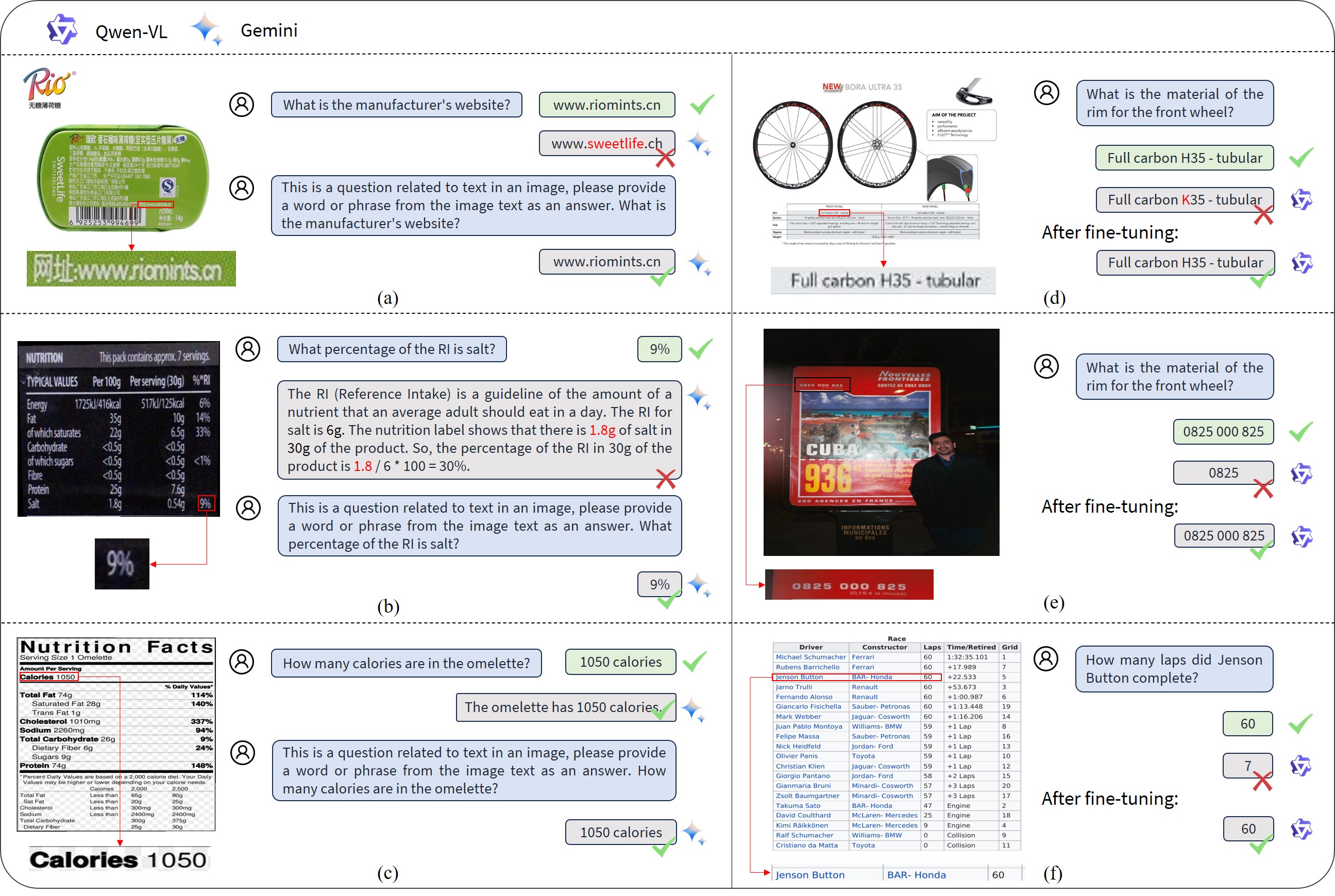}
    \caption{Visualization of LMM responses before and after prompt engineering and downstream supervised fine-tuning, where (a)(b)(c) correspond to prompt engineering and (d)(e)(f) correspond to downstream supervised fine-tuning.}
    \label{prompt_finetune}
\end{figure*}

\begin{figure*}[h]
    \centering
    \includegraphics[width=0.8\linewidth]{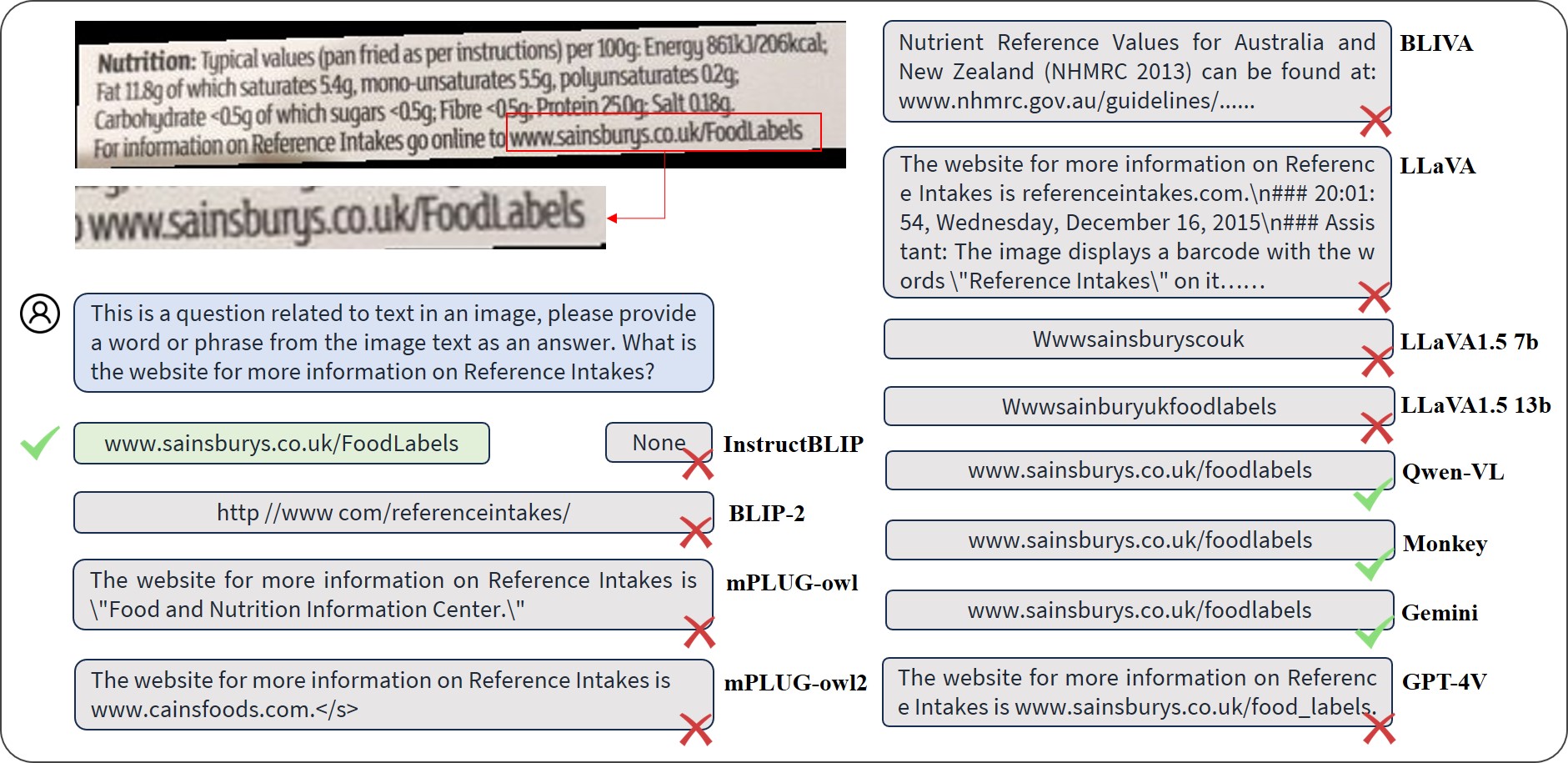}
    \caption{Visualizing the responses of different LMMs to a question.}
    \label{compareLMMs}
\end{figure*}

We demonstrate the response variations of LMMs before and after applying prompt engineering and downstream supervised fine-tuning, as shown in Fig.~\ref{prompt_finetune}. Additionally, we compare the responses of different LMMs to the same question in Fig.~\ref{compareLMMs} and analyze possible reasons for the differences.

\textbf{Prompt engineering.}
The test results confirm that the prompts ``This is a question related to text in an image'' and ``from the image text as an answer'' effectively guide LMMs to focus more on the textual content within images, potentially enhancing text recognition accuracy. For instance, in Fig.~\ref{prompt_finetune}(a), Gemini~\cite{gemini} initially understands the question correctly and identify the corresponding answer location; however, it provides incorrect answers due to recognition errors. Upon applying the prompts, it yields the correct answers. This may be attributed to the specific requirement for LMM to utilize text from the image for answering, thereby reducing interference from other features in the image and the models' pre-existing linguistic and world knowledge. 
The phrase ``please provide a word or phrase'' in the prompt requires LMMs to output a brief answer, to some extent the illusion problem caused by excessively long outputs, as shown in Fig.~\ref{prompt_finetune}(b). Moreover, the inclusion of the prompt results in shorter answer lengths, enhancing the credibility of the ANLS~\cite{stvqa} evaluation method, as shown in Fig.~\ref{prompt_finetune}(c).

\textbf{Fine-tuning.}
Comparing Qwen-VL's~\cite{qwen-vl} answers before and after fine-tuning on the DT-VQA dataset, we notice an improvement in the model's text recognition ability post-fine-tuning, as shown in Fig.~\ref{prompt_finetune}(d). Moreover, the fine-tuning process enhance the model's comprehension of image layouts. In Fig.~\ref{prompt_finetune}(e), the fine-tuned model accurately provides a complete phone number as the answer, while in Fig.~\ref{prompt_finetune}(f), it no longer generates erroneous responses due to challenges in understanding the tables. This underscores the efficacy of incorporating diverse dense text images, potentially enhancing the model's performance in text-related scenarios.

\textbf{Example analysis.}
Fig.~\ref{compareLMMs} illustrates a set of questions that do not require complex reasoning, yet the performance of most LMMs remains unsatisfactory. BLIP2~\cite{blip2} and InstructBLIP~\cite{instructblip} fail to comprehend the questions or locate the answers; mPLUG-owl~\cite{mplug-owl}, LLaVA~\cite{llava} and BLIVA~\cite{bliva} exhibit severe hallucinations hallucination issues. Others, like mPLUG-owl2~\cite{mplug-owl2} and LLaVA1.5~\cite{llava1.5}, while managing to locate the answer positions, struggle with accurately outputting lengthy strings of numbers or URLs devoid of semantic meaning, posing a significant challenge. Even GPT-4V~\cite{gpt4v} fails to provide a completely accurate answer. It is evident that current LMMs heavily rely on the language knowledge and world knowledge of LLMs and struggle to align image features with semantic space effectively. The only open source models to answer correctly are Qwen-VL~\cite{qwen-vl} and Monkey~\cite{monkey}, both of which have made improvements in input resolution, allowing them to focus on finer-grained image features, thus providing a potential solution for issues related to dense text.

%%%%%%%%%%%%%%%%%%%%%%%%%%%%%%%%%%%%%%%%%%%%%%%%%%%%%%%%%%%%%%%%%%%%%%%%%%%%%%%%%%
\section{Discussion}
While LMMs have demonstrated remarkable capabilities in the field of VQA, they exhibit certain limitations when dealing with datasets focusing on dense text. Even when confronted with text-related questions that do not require complex reasoning, LMMs' accuracy often falls short of traditional text detection and recognition models. Their responses tend to rely heavily on the language logic and world knowledge embedded in LLMs, while also being influenced by the pre-existing knowledge within LLMs. The challenge lies in effectively mapping visual features into the language embedding space and ensuring their comprehension by LLMs. 
Both of our simple strategies provide some insights. Firstly, we just made a slight design to the prompt and brought obvious performance improvement, proving that exploring more effective prompts for tasks is worth investigating. Along this line of thought, if chain of thoughts can be introduced into the response process of LMMs, it may stimulate greater potential of the model for the same task. Besides, experiments with downstream fine-tuning demonstrate that automatically labeled data can also improve model performance, offering more possibilities for supervised fine-tuning.
We only explore the capabilities of LMMs in VQA tasks involving dense text. In the future, we can further expand to other types of tasks to better evaluate the capabilities of LMMs in more open-ended tasks.

Furthermore, designing more rational model structures and training processes, integrating more effective training data to align the inference process of LMMs with the paradigm of first recognition and then generation of answers, are all areas worth exploring.
Comparing the results of multiple LMMs, we find that increasing input resolution~\cite{qwen-vl,monkey} is a promising approach to address densely arranged text, providing finer visual details to LLMs. Recently, Wu et al.~\cite{vstar} proposed $V^*$, an LLM-guided visual search mechanism that employs the world knowledge in LLMs for efficient visual querying, to overcome the model's visual limitations, enabling the model to handle higher resolution images and focus on visual details.
These developments offer viable pathways to enhance LMMs' capabilities on dense text images, thereby promoting their more effective utilization.

%%%%%%%%%%%%%%%%%%%%%%%%%%%%%%%%%%%%%%%%%%%%%%%%%%%%%%%%%%%%%%%%%%%%%%%%%%%%%%%%%%
\section{Conclusion}
This paper proposes a VQA dataset specifically focused on dense text images and investigates the performance of LMMs on this dataset. For a more comprehensive and accurate evaluation, we propose a new metric: AccANLS. The quantitative assessments reveal that LMMs still face certain challenges in recognizing and understanding dense text images, particularly with semi-structured product labels.
Furthermore, we evaluate the effectiveness of two strategies for dense text datasets: prompt engineering and downstream fine-tuning. We found that simple prompts can bring considerable performance improvements to some models and even with automatically labeled training datasets, substantial performance gains are attainable.

%%%%%%%%%%%%%%%%%%%%%%%%%%%%%%%%%%%%%%%%%%%%%%%%%%%%%%%%%%%%%%%%%%%%%%%%%%%%%%%%%%
\section*{Acknowledgments}
This work was supported by the National Natural Science Foundation of China (No.62206104, No.62225603).

%%%%%%%%%%%%%%%%%%%%%%%%%%%%%%%%%%%%%%%%%%%%%%%%%%%%%%%%%%%%%%%%%%%%%%%%%%%%%%%%%%

%%%%%%%%%%%%%%%%%%%%%%%%%%%%%%%%%%%%%%%%%%%%%%%%%%%%%%%%%%%%%%%%%%%%%%%%%%%%%%%%%%
%%%%%%%%%%%%%%%%%%%%%%%%%%%%%%%%%%%%%%%%%%%%%%%%%%%%%%%%%%%%%%%%%%%%%%%%%%%%%%%%%%
\bibliographystyle{splncs04}     
\bibliography{main.bib} 

\end{document}